\documentclass[a4paper,twocolumn,10pt]{article}
\usepackage{eurosis}

\usepackage{url}
\usepackage{natbib}
\usepackage[T1]{fontenc}  
\usepackage{graphicx}
\usepackage{array}
\usepackage{algorithm,algpseudocode}
\usepackage{amsmath}
\usepackage{amsfonts}
\usepackage{amssymb}
\usepackage[dvipsnames]{xcolor}
\usepackage{changes}

\bibpunct[; ]{(}{)}{,}{a}{}{;}

\usepackage{graphicx}
\usepackage{array, makecell} 
\usepackage{enumerate}   
\usepackage{changes}

\definechangesauthor[color=NavyBlue]{LR}
\definechangesauthor[color=blue]{MS}
\title{\LARGE \bf
On the recognition of the game type based on physiological signals and eye tracking}

\author{Łukasz Czekaj$^{1}$, Łukasz Radzinski$^{1}$, Mateusz Kolimaga$^{1}$, Jakub Domaszewicz$^{1}$, \\
Robert Kitłowski$^{1}$, Mariusz Szwoch$^{2}$, Włodzisław Duch$^{3}$
\thanks{$^{1}$Aidmed, https://www.aidmed.ai/, Poland}%
\thanks{$^{2}$Faculty of Electronics, Telecommunications and Informatics, Gdańsk University of Technology, Poland}
\thanks{$^{3}$Dept. of Informatics, Institute of Engineering and Technology, Faculty of Physics, Astronomy \& Informatics, Nicolaus Copernicus University, Poland}
}

\newcommand{\reffig}[1]{FIG.~\ref{#1}}
\newcommand{\reftab}[1]{TAB.~\ref{#1}}

\begin{document}

\maketitle
\thispagestyle{empty}
\pagestyle{empty}
\begin{abstract}
%
Automated interpretation of signals 
yields many impressive applications from the area of affective computing and human activity recognition (HAR). In this paper we ask the question about possibility of cognitive activity recognition on the base of particular set of signals. 
We use recognition of the game played by the participant as a playground for exploration of the problem. We build classifier of three different games (Space Invaders, Tetris, Tower Defence) and inter-game pause. We validate classifier in the player-independent and player-dependent scenario. We discuss the improvement in the player-dependent scenario in the context of biometric person recognition.
On the base of the results obtained in game classification, we consider potential applications in smart surveillance and quantified self.
\end{abstract}
\section{INTRODUCTION}
Easy access to the wide spectrum of wearable and mobile sensors drives interest of researchers toward automated and continuous recognition of various human activities. Among the motivations are: activity tracking in such areas as healthcare and elder care, smart surveillance, intelligent environment, education, entertainment and virtual reality, biofeedback and wellness (emotional strain and stress assessment).
Implementations of research results together with Big Data technologies and individual engaged in the self-tracking manifest as cultural phenomenon of {\it quantifies self} movement
\cite{swan2013quantified}

{\it Affective computing} is fast developing multidisciplinary research area bringing together artificial intelligence, cognitive and social sciences, with the aim of detection, recognition and interpretation of human affects \cite{tao2005affective, poria2017review, WANG202219}.
Affective computing utilize various machine learning techniques in processing data acquired from wearable and mobile sensors, cameras and other dedicated input devices.
Among its interest are: automated emotion recognition \cite{s20030592}, 
affective gaming \cite{hudlicka2008affective} (smart games and player engagement),
stress detection and mental workload \cite{stress_1}.

Closely related is so called mind-reading with impressive results in recognition of cognitive states  from fMRI data
\cite{norman2006beyond} 
and generating images from EEG signals \cite{bai2023dreamdiffusion}.

We were inspired by various results in quantified self \cite{swan2012sensor}, HAR
\cite{HAR_review_2}, 
mental workload and time pressure detection \cite{workload_0}, games as affective stimulation \cite{games_as_affective_stimuli}. 

In this paper we focus on the problem of the recognition of cognitive activity, i.e. classification of what type of cognitive task from a given set is performed by the person at a given moment. The input to the classifier are physiological signals along with eye tracking data (for brevity we call them signals). We simplify this general problem and study it as a task of distinguishing between different game types and inter-game pause (for short we refer to this pause extended game type classification as game classification). Our work aligns with the framework of sensor-based, single-user activity recognition
and the results may be viewed as extension of classical HAR from physical to mental activities.

The intuitions behind presented research are as follows. Despite the fact that the participant is sitting in front of the computer screen throughout the game, and the physical conditions do not change much between different games, the virtual environment changes a lot. Different games engage participant in different ways. Theirs dynamics, speed, turns design and time pressure, and overall concentration level required by the game reflect in signals recorded during the game. Game specific screen layout is correlated with eye-movement patterns.
The mentioned factors enable identification of the type of game. In this paper we present XGBoost~\cite{XGBoost} based detector (multi-label classifier) fuelled by the features obtained from the following signals: electrocardiography (ECG), respiration (RESP), galvanic skin response (GSR) and eye tracking (EYETR).
All those signals can be recorded in non-invasive way and the applications using such  signals do not require professional training. Moreover, the sensors may be implemented in non-distributive way without need of wearables (e.g. ECG recording from the steering  \cite{s22114198}). 

\begin{figure}
\includegraphics[width=8.5cm]{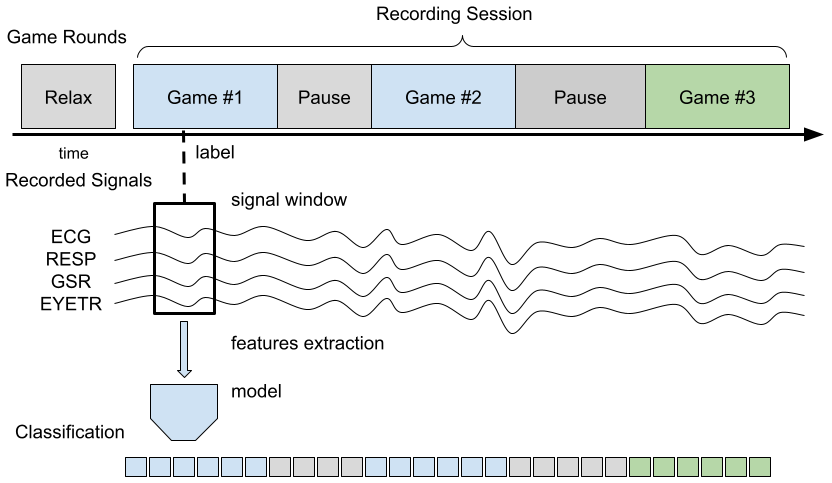}
\caption{{\it Data collection and signal processing.} The top of the picture represents a recording session: a player plays multiple rounds of games (round is depicted as a box, different colours denote different games) separated by pauses (gray boxes). The recording session is preceded with eye tracker calibration and short relaxation. During the session, signals (ECG, RESP, GSR, EYETR) are recorded along with information about game type or a pause. Sliding window of the width $15$\,s and step $1$\,s is applied on the signals and is used for feature extraction. Window is labelled according to its centre. Subject normalized features vector is passed to the multi-class model. In the prediction mode, for every data window, models provide 4 probabilities: 1 for each game type/pause. Then the label corresponding to the class with the highest probability value is taken.
\label{pic:signal_pipeline}}
\end{figure}

\begin{figure}
\includegraphics[width=7cm]{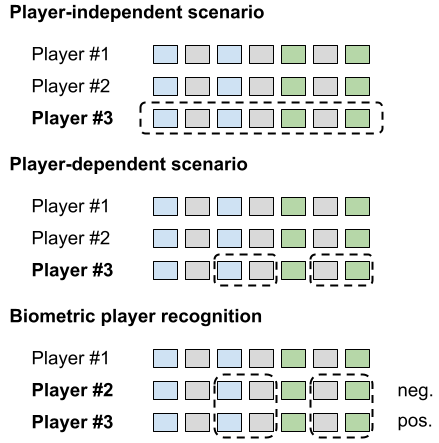}
\caption{{\it Models and Evaluation Scenarios.} 
The top two panels refer to game classification tasks, the bottom one refers to player recognition. 
Leave-one-out cross validation is used for evaluation of game classification: for each LOOCV iteration, one player is selected as a test player (in bold); in {\it player-independent scenario} - recordings for test player are completely excluded from the train dataset;  in {\it player-dependent scenario} - recordings from the test player are split between train and test datasets.
In {\it biometric player recognition} one test player is selected as {\it positive} and one as {\it negative}. Data from the rest of the players form negative examples in train dataset.
Row depicts recording session for player; colour boxes represent game rounds (different colours denotes different games); gray boxes represent inter-game pauses. Dashed lines denote test data and the complement forms training data.
\label{pic:evaluation}}
\end{figure}

We build two types of game detectors: player-independent and player-dependent. Detectors are evaluated using leave-one-out cross-validation (LOOCV) - one test player is selected in each iteration. For player-independent detector, recordings for test player are completely excluded from the train dataset. For player-dependent detectors, recordings from the test player are split between train and test datasets. The model also learns features specific for the test player (see~\reffig{pic:evaluation} for details). Since we observed better performance of the player-dependent detector, we asked if that difference should be attributed to the {\it player's related clusterisation} of recorded signals (e.g. player related shift of mean HR) or player's preferences of the {\it specific game playing style} (i.e. clustering of signals recorded for given game which may be interpreted as playing style). To answer the question we build a biometric player recognition model 
\cite{odinaka2012ecg} 
that predicted if a given part of signals belongs to the test user. Our results suggest that {\it player's related clustering} is important factor in model performance improvement.

\section{METHODS}

\subsection{Data collection}
The data collection process is depicted in~\reffig{pic:signal_pipeline}. Each player participated in one recording session. The recording session was preceded with eye tracker calibration and a short relaxation (for the purpose of heart rate and breath rate stabilisation). The recording session consisted of game rounds interwoven with pauses. Game round took $\sim 5$\,mins and pause took $\sim 1$\,min. There were $4$ consecutive rounds for each game. Participants played in $3$ games: Space Invaders, Tetris and Tower Defense. Games differ in graphical layout and dynamics: (i) Space Invaders is characterized by spawning waves of enemies; (ii) in Tetris the gameplay speeds up and the time pressure increases at the end of each round; (iii) Tower Defense is a sub-genre of turn-based strategy game. 
Data collection during recording session was performed with following devices: ECG and respiration (bioimpedance method) were recorded with the Aidmed One recorder (\cite{czekaj2020validation}), GSR was recorded with Shimmer device (https://shimmersensing.com/), eye positions were collected with Talon (https://talonvoice.com/) and Tobii Eye Tracker (https://gaming.tobii.com/). 
Data was collected from $20$ volunteers in age 20-40 years, 5 female. 
All the data was collected in the same laboratory station. The participants were instructed about data collection protocol. Desktop application developed in Python was used to control the protocol. Data were collected in a separate room in a quiet, isolated environment.

\subsection{Detector}
In this paper we present results for 3 classification tasks: game classification for i) player-independent and ii) player-dependent scenario, iii) biometric player recognition. For each task we built XGBoost based model fed by features extracted from the signals (see~\reffig{pic:signal_pipeline}). For tasks i) and ii) we developed multiclass models (multi:softmax) with game type including  pause as a label. For task iii) we used binary loss (binary:logistic) and label denotes selected player. 
Biometric player recognition mimics player-dependent scenario in the way of data split, however the goal is to recognize player on the base of signals sample. 
We applied class balancing with example weighting in all models.
For details of train and test datasets preparation see~\reffig{pic:evaluation}.

For all tasks we build separate models for 3 sets of features derived from increasing number of signals: I) SIG-1: ECG, RESP; II) SIG-2: ECG, RESP, GSR; III) SIG-3: ECG, RESP, GSR, EYETR.
Our motivation here was to evaluate a hierarchy of signals/sensors from less to more intrusive: Aidmed One recorder is chest belt wearable, GSR requires fingers attached electrodes and eye tracking requires special camera placed in front of player or the use of special glasses. Signal processing and feature extraction pipeline was the same in all models.

Based on ECG, we calculated heart rate variability (HRV) features using (https://github.com/Aura-healthcare) \cite{hrv_1}.
GSR signal was decomposed into phasic and tonic components and analysed with Neurokit \cite{Makowski2021neurokit} toolkit. 
A detailed list of features is presented below.

ECG based features:
\begin{itemize}
    \item {\it AVG\_HR\_15s} - average HR from 15\,s window 
    \item {\it AVG\_HR\_5s} - average HR from central 5\,s range  
    \item {\it AVG\_HR\_RATIO} - ratio of {\it AVG\_HR\_5s}-to-{\it AVG\_HR\_15s}  
    \item {\it AVG\_HR\_DIFF} - difference between average HR of left and right half of the signal window 
    \item {\it SDNN} - standard deviation of inter-beat intervals 
    \item {\it RMSSD} - root-mean-square of successive differences 
    \item {\it LF\_POWER} - relative power of the low-frequency band ($\leq 0.15$\,Hz) 
    \item {\it HF\_POWER} - relative power of the low-frequency band ($0.15-0.4$\,Hz) 
    \item {\it LF\_HF\_RATIO} - ratio of LF-to-HF power
    \item {\it TOTAL\_POWER} - total power of the heart rate signal 
\end{itemize}
Respiration based features:
\begin{itemize}
    \item {\it RESP\_AMP} - respiration amplitude
    \item {\it RESP\_DOM\_FREQ} - dominant frequency of respiration signal (estimation of respiration rate)
    \item {\it RESP\_PEAKS} - peaks number in respiration signal (estimation of respiration rate)
\end{itemize}
GSR based features (features are calculated separately for the tonic and phasic components):
\begin{itemize}
    \item {\it AVG\_GSR\_15s} - average value of GSR component from 15\,s window
    \item {\it AVG\_GSR\_5s} - average value of GSR component from central 5\,s range
    \item {\it AVG\_GSR\_RATIO} - ratio of {\it AVG\_GSR\_5s}-to-{\it AVG\_GSR\_15s}
    \item {\it AVG\_MEAN\_GSR\_DIFF} - difference between average GSR component of left and right half of the signal window
    \item {\it STD\_GSR} - standard deviation of GSR signal component
    \item {\it GSR\_PEAKS} - number of peaks in GSR (raw signal)
\end{itemize}
Eye tracking based features (gaze position in the screen coordinate,  gaze outside the screen was filtered out):
\begin{itemize}
    \item {\it BLINKS} - blinks number
    \item {\it GAZE\_OUTSIDE} - pc. of time when gaze was not located on screen
    \item {\it AVG\_POSITION} - average of eye position
    \item {\it STD\_POSITION} - standard deviation of gaze position
    \item {\it KURT\_POSITION} - kurtosis of gaze position
    \item {\it AVG\_VELOCITY} - average of gaze position difference
    \item {\it STD\_VELOCITY} - standard deviation of gaze position difference
    \item {\it KURT\_VELOCITY} - kurtosis of gaze position difference    
\end{itemize}
Computations were performed using Python 3.8 \cite{python_ref}. 

\section{EXPERIMENTS AND RESULTS}
\begin{figure}
\includegraphics[width=8cm]{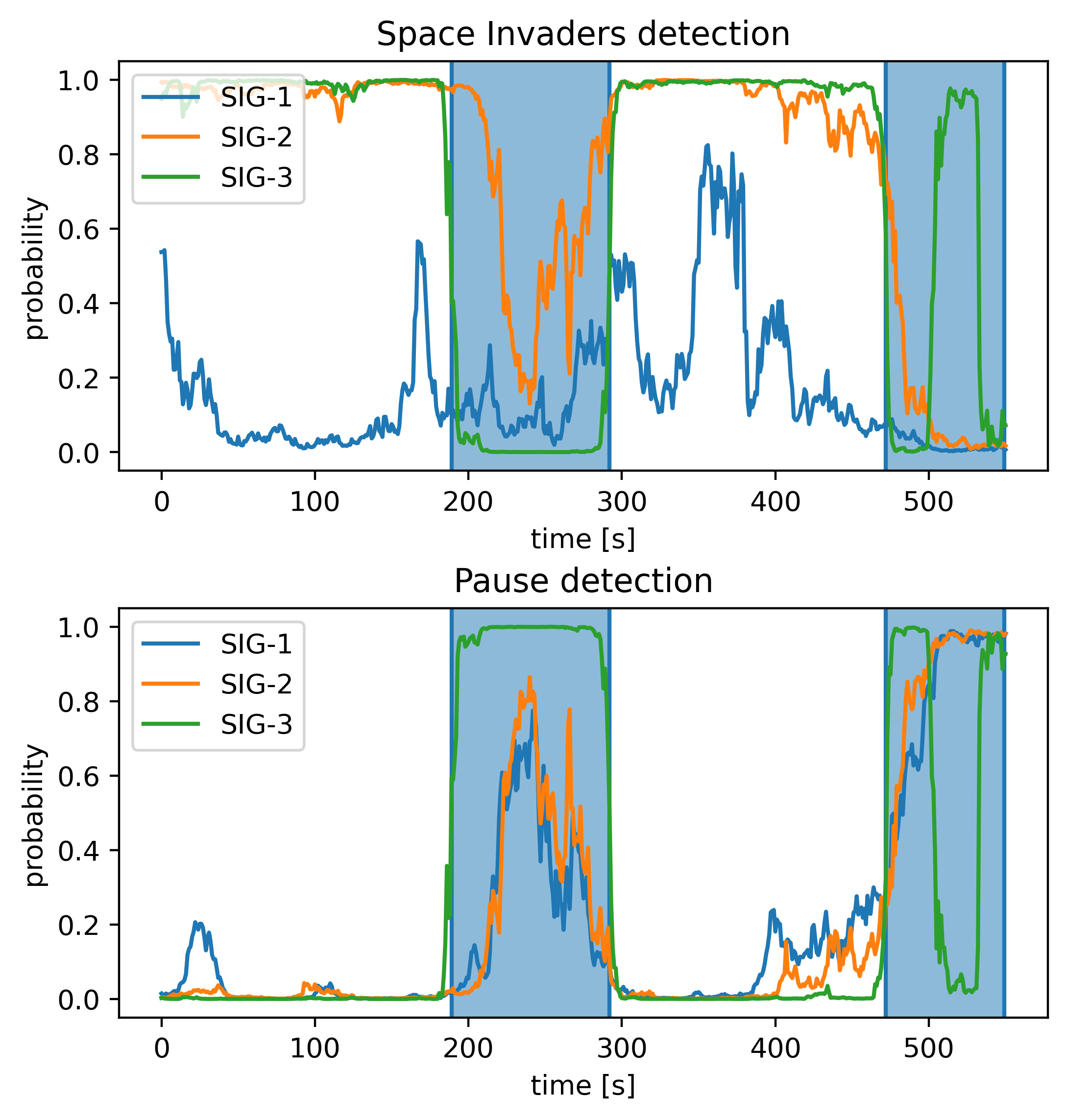}
\caption{{\it Detector output for player-dependent scenario} Figure presents fragment of recording sessions containing Space Invaders and pauses (blue rectangles). Top and bottom panels show output of Space Invaders and pause detector, respectively. Please note that for clarity we only plot the probabilities of 2 from 4 class. Colour lines represent result for different set of signals (SIG-1: ECG, RESP; SIG-2: ECG, RESP, GSR; SIG-3: ECG, RESP, GSR, EYETR).
\label{pic:signals_evaluation}}
\end{figure}

Results for the player-independent classification, player-dependent classification and biometric player recognition are summarized in  \reftab{tab:evaluation_independent}, \reftab{tab:evaluation_calibrated} 
and \reftab{tab:evaluation} respectively. Raw output of the models is presented in~\reffig{pic:signals_evaluation}.
For each experiment we present results of random model (random) as a baseline and then results for selected sets of signals: SIG-1, SIG-2, SIG-3. 
For each experiment we reported accuracy, precision, recall and F1 score. Data from different games and cross validation iterations were aggregated using multiclass macro averaging and balanced accuracy score \cite{mosley2013balanced}.

All game classification models significantly outperform baseline random classifier - even simple ECG and respiration based one. Adding GSR features do not improve player-independent classifier, but they help in the player-dependent classifier. Adding features extracted from eye tracking help a lot in both game classification scenarios, but do not improve player recognition much. All sets of signals carry information about player identity. Our interpretation of the results is that the ECG, respiration and GSR features are more specific to the user and the technologies for cognitive activity recognition based on that signals will require user specific calibration of the model. On the other hand eye tracking carry information about game specific layout and the reaction on visual signals generalize between users.

\begin{table}
\caption{Player-independent classification.}
\label{tab:evaluation_independent}
\centering
\begin{tabular}{ |c|r|r|r|r| } 
signals & acc. & prec. & rec. & F1 \\
\hline
random & 0.25 & 0.25 & 0.25 & 0.25 \\
\hline
SIG-1 &0.50 & 0.46 & 0.48 & 0.47 \\
SIG-2 & 0.52 & 0.48 & 0.49 & 0.48 \\
SIG-3 & 0.71& 0.68& 0.70& 0.69
\end{tabular}
\end{table}

\begin{table}
\caption{Player-dependent classification.}
\label{tab:evaluation_calibrated}
\centering
\begin{tabular}{ |c|r|r|r|r| } 
signals & acc. & prec. & rec. & F1 \\
\hline
random & 0.25 & 0.25 & 0.25 & 0.25 \\
\hline
SIG-1 &0.62 & 0.58 & 0.58 & 0.58 \\
SIG-2 & 0.76 & 0.75 & 0.74 & 0.75 \\
SIG-3 & 0.89& 0.88& 0.89& 0.89
\end{tabular}
\end{table}

\begin{table}
\caption{Biometric player recognition}
\label{tab:evaluation}
\centering
\begin{tabular}{ |c|r|r|r|r| } 
signals & acc. & prec. & rec. & F1 \\
\hline
random & 0.5 & 0.5 & 0.5 & 0.5 \\
\hline
SIG-1 & 0.78 & 0.82 & 0.78 & 0.78 \\
SIG-2 & 0.84 &  0.88 & 0.84 & 0.84\\
SIG-3 & 0.87 & 0.90 & 0.87 & 0.87 
\end{tabular}
\end{table}

\section{CONCLUSIONS}
Using the example of game type/pause classification during play we have demonstrated the possibility of cognitive activity recognition. We have discovered the user specific characteristic of the heart rate, respiration and GSR signal. Our results may find applications in tasks related with smart surveillance (e.g. in profession like flight control where mental overload/fatigue may be crucial factor for safety and performance), affect-aware video games and smart games (e.g. feedback provided to student about focus) and e-learning software, and new methods for human-machine interactions. 
Cognitive activity recognition implemented as a part of quantified self
application may be used as qualitative feedback loops for behavior change.

Since each player participated in only one recording session, more research is required to assess the necessity of detector calibration between recording sessions and how it is related to the detector calibration for the player.

\section*{ACKNOWLEDGMENT}

This work is a part of the project `Development of Software Development Kit for utilizing biosignals from wearable sensor to improve user's interaction in gaming`, financed by the  National Centre for Research and Development (NCBiR), Poland, under the agreement POIR.01.02.00-00-0212. We would like to thank PICTEC team and AIDMED team for their involvement in this project.

\bibliography{bibliography}

\begin{thebibliography}{21}
\providecommand{\natexlab}[1]{#1}
\providecommand{\url}[1]{\texttt{#1}}
\providecommand{\urlprefix}{URL }
\expandafter\ifx\csname urlstyle\endcsname\relax
  \providecommand{\doi}[1]{doi:\discretionary{}{}{}#1}\else
  \providecommand{\doi}{doi:\discretionary{}{}{}\begingroup
  \urlstyle{rm}\Url}\fi

\bibitem[{Bai et~al.(2023)Bai, Wang, Cao, Ge, Yuan, and
  Shan}]{bai2023dreamdiffusion}
Bai Y.; Wang X.; Cao Y.; Ge Y.; Yuan C.; and Shan Y., 2023.
\newblock \emph{DreamDiffusion: Generating High-Quality Images from Brain EEG
  Signals}.
\newblock \emph{arXiv preprint arXiv:230616934}.

\bibitem[{Bouchabou et~al.(2021)Bouchabou, Nguyen, Lohr, LeDuc, and
  Kanellos}]{HAR_review_2}
Bouchabou D.; Nguyen S.M.; Lohr C.; LeDuc B.; and Kanellos I., 2021.
\newblock \emph{A Survey of Human Activity Recognition in Smart Homes Based on
  IoT Sensors Algorithms: Taxonomies, Challenges, and Opportunities with Deep
  Learning}.
\newblock \emph{Sensors}, 21, no.~18.

\bibitem[{Chen and Guestrin(2016)}]{XGBoost}
Chen T. and Guestrin C., 2016.
\newblock \emph{{XGBoost}: A Scalable Tree Boosting System}.
\newblock In \emph{Proc. of the 22nd ACM SIGKDD Int. Conf. on Knowledge
  Discovery and Data Mining}. 785--794.

\bibitem[{Chen et~al.(2019)Chen, Hsiao, Zheng, Lee, and
  Lin}]{games_as_affective_stimuli}
Chen Y.C.; Hsiao C.C.; Zheng W.D.; Lee R.G.; and Lin R., 2019.
\newblock \emph{Artificial neural networks-based classification of emotions
  using wristband heart rate monitor data}.
\newblock \emph{Medicine}, 98, no.~33.

\bibitem[{Czekaj et~al.(2020)Czekaj, Domaszewicz, Radzinski, Jarynowski,
  Kitlowski, and Doboszynska}]{czekaj2020validation}
Czekaj L.; Domaszewicz J.; Radzinski L.; Jarynowski A.; Kitlowski R.; and
  Doboszynska A., 2020.
\newblock \emph{Validation and usability of AIDMED-telemedical system for
  cardiological and pulmonary diseases}.
\newblock \emph{E-methodology}, 7, no.~7, 125--139.

\bibitem[{Dzedzickis et~al.(2020)Dzedzickis, Kaklauskas, and
  Bucinskas}]{s20030592}
Dzedzickis A.; Kaklauskas A.; and Bucinskas V., 2020.
\newblock \emph{Human Emotion Recognition: Review of Sensors and Methods}.
\newblock \emph{Sensors}, 20, no.~3.

\bibitem[{Giannakakis et~al.(2022)Giannakakis, Grigoriadis, Giannakaki,
  Simantiraki, Roniotis, and Tsiknakis}]{stress_1}
Giannakakis G.; Grigoriadis D.; Giannakaki K.; Simantiraki O.; Roniotis A.; and
  Tsiknakis M., 2022.
\newblock \emph{Review on Psychological Stress Detection Using Biosignals}.
\newblock \emph{IEEE Transactions on Affective Computing}, 13, no.~1, 440--460.

\bibitem[{Hudlicka(2008)}]{hudlicka2008affective}
Hudlicka E., 2008.
\newblock \emph{Affective computing for game design}.
\newblock In \emph{Proceedings of the 4th Intl. North American Conference on
  Intelligent Games and Simulation}. McGill University Montreal, 5--12.

\bibitem[{Makowski et~al.(2021)Makowski, Pham, Lau, Brammer, Lespinasse, Pham,
  Schölzel, and Chen}]{Makowski2021neurokit}
Makowski D.; Pham T.; Lau Z.J.; Brammer J.C.; Lespinasse F.; Pham H.; Schölzel
  C.; and Chen S.H.A., 2021.
\newblock \emph{{NeuroKit}2: A Python toolbox for neurophysiological signal
  processing}.
\newblock \emph{Behavior Research Methods}, 53, no.~4, 1689--1696.

\bibitem[{Mosley(2013)}]{mosley2013balanced}
Mosley L., 2013.
\newblock \emph{A balanced approach to the multi-class imbalance problem}.
\newblock \emph{Iowa State University Digital Repository}.

\bibitem[{Nickel and Nachreiner(2003)}]{workload_0}
Nickel P. and Nachreiner F., 2003.
\newblock \emph{Sensitivity and Diagnosticity of the 0.1-Hz Component of Heart
  Rate Variability as an Indicator of Mental Workload}.
\newblock \emph{Human Factors}, 45, no.~4, 575--590.

\bibitem[{Norman et~al.(2006)Norman, Polyn, Detre, and
  Haxby}]{norman2006beyond}
Norman K.A.; Polyn S.M.; Detre G.J.; and Haxby J.V., 2006.
\newblock \emph{Beyond mind-reading: multi-voxel pattern analysis of fMRI
  data}.
\newblock \emph{Trends in cognitive sciences}, 10, no.~9, 424--430.

\bibitem[{Odinaka et~al.(2012)Odinaka, Lai, Kaplan, O'Sullivan, Sirevaag, and
  Rohrbaugh}]{odinaka2012ecg}
Odinaka I.; Lai P.H.; Kaplan A.D.; O'Sullivan J.A.; Sirevaag E.J.; and
  Rohrbaugh J.W., 2012.
\newblock \emph{ECG biometric recognition: A comparative analysis}.
\newblock \emph{IEEE Transactions on Information Forensics and Security}, 7,
  no.~6, 1812--1824.

\bibitem[{Pham et~al.(2021)Pham, Lau, Chen, and Makowski}]{hrv_1}
Pham T.; Lau Z.J.; Chen S.H.A.; and Makowski D., 2021.
\newblock \emph{Heart Rate Variability in Psychology: A Review of HRV Indices
  and an Analysis Tutorial}.
\newblock \emph{Sensors}, 21, no.~12.

\bibitem[{Poria et~al.(2017)Poria, Cambria, Bajpai, and
  Hussain}]{poria2017review}
Poria S.; Cambria E.; Bajpai R.; and Hussain A., 2017.
\newblock \emph{A review of affective computing: From unimodal analysis to
  multimodal fusion}.
\newblock \emph{Information fusion}, 37, 98--125.

\bibitem[{Swan(2012)}]{swan2012sensor}
Swan M., 2012.
\newblock \emph{Sensor mania! the internet of things, wearable computing,
  objective metrics, and the quantified self 2.0}.
\newblock \emph{Journal of Sensor and Actuator networks}, 1, no.~3, 217--253.

\bibitem[{Swan(2013)}]{swan2013quantified}
Swan M., 2013.
\newblock \emph{The quantified self: Fundamental disruption in big data science
  and biological discovery}.
\newblock \emph{Big data}, 1, no.~2, 85--99.

\bibitem[{Tao and Tan(2005)}]{tao2005affective}
Tao J. and Tan T., 2005.
\newblock \emph{Affective computing: A review}.
\newblock In \emph{International Conference on Affective computing and
  intelligent interaction}. Springer, 981--995.

\bibitem[{Van~Rossum and Drake(2009)}]{python_ref}
Van~Rossum G. and Drake F.L., 2009.
\newblock \emph{Python 3 Reference Manual}.
\newblock CreateSpace, Scotts Valley, CA.

\bibitem[{Wang et~al.(2022)Wang, Song, Tao, Liotta, Yang, Li, Gao, Sun, Ge,
  Zhang, and Zhang}]{WANG202219}
Wang Y.; Song W.; Tao W.; Liotta A.; Yang D.; Li X.; Gao S.; Sun Y.; Ge W.;
  Zhang W.; and Zhang W., 2022.
\newblock \emph{A systematic review on affective computing: emotion models,
  databases, and recent advances}.
\newblock \emph{Information Fusion}, 83-84, 19--52.

\bibitem[{Warnecke et~al.(2022)Warnecke, Ganapathy, Koch, Dietzel, Flormann,
  Henze, and Deserno}]{s22114198}
Warnecke J.M.; Ganapathy N.; Koch E.; Dietzel A.; Flormann M.; Henze R.; and
  Deserno T.M., 2022.
\newblock \emph{Printed and Flexible ECG Electrodes Attached to the Steering
  Wheel for Continuous Health Monitoring during Driving}.
\newblock \emph{Sensors}, 22, no.~11.

\end{thebibliography}

\end{document}